%% file: main.tex
\documentclass[10pt,twocolumn,letterpaper]{article}

\usepackage{wacv}              % To produce the CAMERA-READY version
\input{preamble}

\definecolor{wacvblue}{rgb}{0.21,0.49,0.74}
\usepackage[pagebackref,breaklinks,colorlinks,allcolors=wacvblue]{hyperref}
\usepackage{orcidlink}  % Needs to be loaded after hyperref to prevent option clashes

\def\wacvPaperID{544} % *** Enter the WACV Paper ID here
\def\confName{WACV}
\def\confYear{2026}

\title{Correcting and Quantifying Systematic Errors in 3D Box Annotations for Autonomous Driving}

\author{
Alexandre Justo Miro\orcidlink{0009-0009-5436-097X}\textsuperscript{1,2},
Ludvig af Klinteberg\orcidlink{0000-0001-7425-8029}\textsuperscript{2},
Bogdan Timus\textsuperscript{1},
Aron Asefaw\textsuperscript{3}, \\
Ajinkya Khoche\textsuperscript{1,3},
Thomas Gustafsson\textsuperscript{1},
Sina Sharif Mansouri\textsuperscript{1},
Masoud Daneshtalab\textsuperscript{2} \\
\textsuperscript{1} Traton Group R\&D, \textsuperscript{2} Mälardalen University, \textsuperscript{3} KTH Royal Institute of Technology \\
\textsuperscript{1} {\tt\small \{alexandre.justo.miro, bogdan.timus, ajinkya.khoche, thomas.gustafsson, } \\
{\tt\small sina.sharif.mansouri\}@se.traton.com}, \textsuperscript{2} {\tt\small \{ludvig.af.klinteberg, masoud.daneshtalab\}@mdu.se}, \\
\textsuperscript{3} {\tt\small asefaw@kth.se}
}

\begin{document}
\maketitle
\input{0_abstract}    
\input{1_introduction}
\input{2_related_work}
\input{3_methodology}
\input{4_experiments}
\input{5_conclusion_and_future_work}
\input{6_acknowledgments}
%\pagebreak  % Force references in a separate page at the end because they are not included in the page limit count
{
    \small
    \bibliographystyle{ieeenat_fullname}
    \bibliography{main}
}

\end{document}

%% file: preamble.tex
\usepackage{graphicx}
\usepackage{amsmath}
\usepackage{amssymb}
\usepackage{booktabs}
\usepackage{bm}

\usepackage[acronym,nomain,nonumberlist]{glossaries}
\usepackage{acronym}

\usepackage{plotting_preamble}
\usepackage{datatool} % For reading CSV files

\usepackage{makecell}

\usepackage{siunitx}

\usepackage{nicefrac}

\usepackage{amsfonts}
\usepackage{algorithmic}
\usepackage{textcomp}
\usepackage{subcaption}
\usepackage{comment}
\usepackage{standalone}

\DeclareMathOperator*{\argmin}{arg\,min}

\newacronym{radar}{RADAR}{Radio Detection and Ranging}
\newacronym{soa}{SoA}{State of the Art}
\newacronym{sota}{SOTA}{state-of-the-art}
\newacronym{lidar}{LiDAR}{Light Detection and Ranging}
\newacronym{ps}{PS}{Pattern Search}
\newacronym{ctra}{CTRA}{Constant Turn Rate and Acceleration}
\newacronym{zod}{ZOD}{Zenseact Open Dataset}
\newacronym{ipd}{IPD}{Inlier Points Difference}
\newacronym{ede}{EDE}{Euclidean Distance Error}
\newacronym{dede}{DEDE}{Decoupled Euclidean Distance Error}
\newacronym{sede}{SEDE}{Spread of Euclidean Distance Error}
\newacronym{sdede}{SDEDE}{Spread of Decoupled Euclidean Distance Error}

\usepackage{array}
\newcommand{\PreserveBackslash}[1]{\let\temp=\\#1\let\\=\temp}
\newcolumntype{C}[1]{>{\PreserveBackslash\centering}p{#1}}
\newcolumntype{R}[1]{>{\PreserveBackslash\raggedleft}p{#1}}
\newcolumntype{L}[1]{>{\PreserveBackslash\raggedright}p{#1}}

%% file: 0_abstract.tex
\begin{abstract}
Accurate ground truth annotations are critical to supervised learning and evaluating the performance of autonomous vehicle systems. These vehicles are typically equipped with active sensors, such as LiDAR, which scan the environment in predefined patterns. 3D box annotation based on data from such sensors is challenging in dynamic scenarios, where objects are observed at different timestamps, hence different positions. Without proper handling of this phenomenon, systematic errors are prone to being introduced in the box annotations.
Our work is the first to discover such annotation errors in widely used, publicly available datasets. Through our novel offline estimation method, we correct the annotations so that they follow physically feasible trajectories and achieve spatial and temporal consistency with the sensor data.
For the first time, we define metrics for this problem; and we evaluate our method on the Argoverse 2, MAN TruckScenes, and our proprietary datasets. Our approach increases the quality of box annotations by more than 17\% in these datasets. Furthermore, we quantify the annotation errors in them and find that the original annotations are misplaced by up to \qty{2.5}{\meter}, with highly dynamic objects being the most affected. Finally, we test the impact of the errors in benchmarking and find that the impact is larger than the improvements that state-of-the-art methods typically achieve \wrt the previous state-of-the-art methods; showing that accurate annotations are essential for correct interpretation of performance. Our code is available at \href{https://github.com/alexandre-justo-miro/annotation-correction-3D-boxes}{https://github.com/alexandre-justo-miro/annotation-correction-3D-boxes}.
\end{abstract}
\glsresetall

%% file: 1_introduction.tex
\section{Introduction} \label{sec:introduction}

Accurate ground truth annotations are a cornerstone in the development, evaluation, and testing of autonomous driving systems. Modern perception pipelines are often powered by deep neural networks trained in a supervised manner and so rely on high-quality labeled data. Annotations are estimations of ground truth, and can be obtained in different ways, \eg manually by humans, via an automatic labeling method, or a combination. The error of these estimations in autonomous driving datasets is widely assumed to be orders of magnitude below the estimation error of the downstream tasks that are benchmarked on the annotations. However, this assumption is seldom revisited, and the accuracy of annotations and their resemblance to the ground truth have been highly overlooked.

\begin{figure*}[htbp]
\centering
    \begin{tabular}{cc}
        \begin{tabular}{c}

            % CAMERA IMAGES - 3 DATASETS
            \begin{subfigure}[c]{0.9\textwidth}
                \begin{tabular}{p{0.3\textwidth}p{0.3\textwidth}p{0.3\textwidth}}
                    \includegraphics[trim={10cm 2cm 8cm 19cm},clip,width=0.3\textwidth]{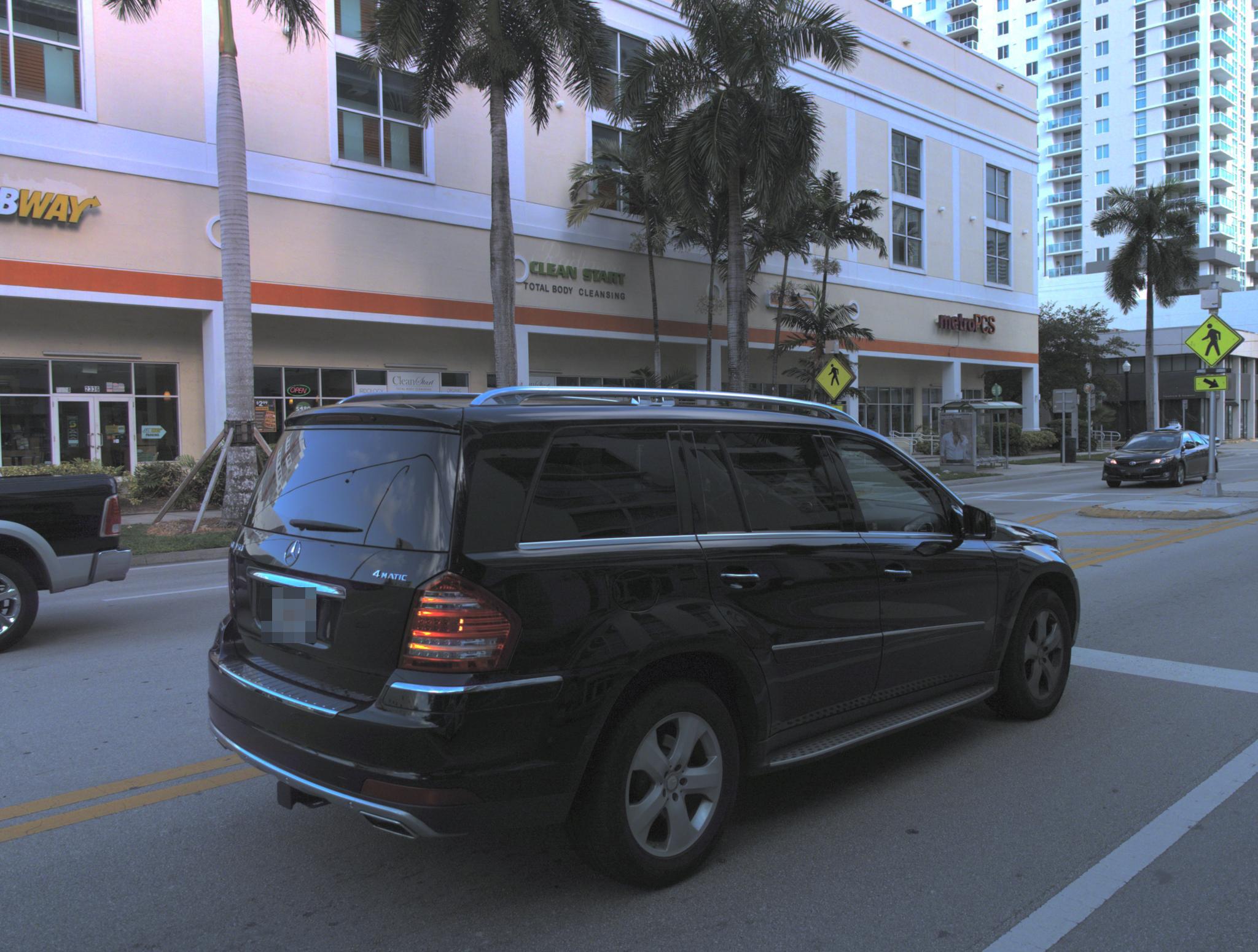}&
                    \includegraphics[trim={9cm 1cm 39cm 18.5cm},clip,width=0.3\textwidth]{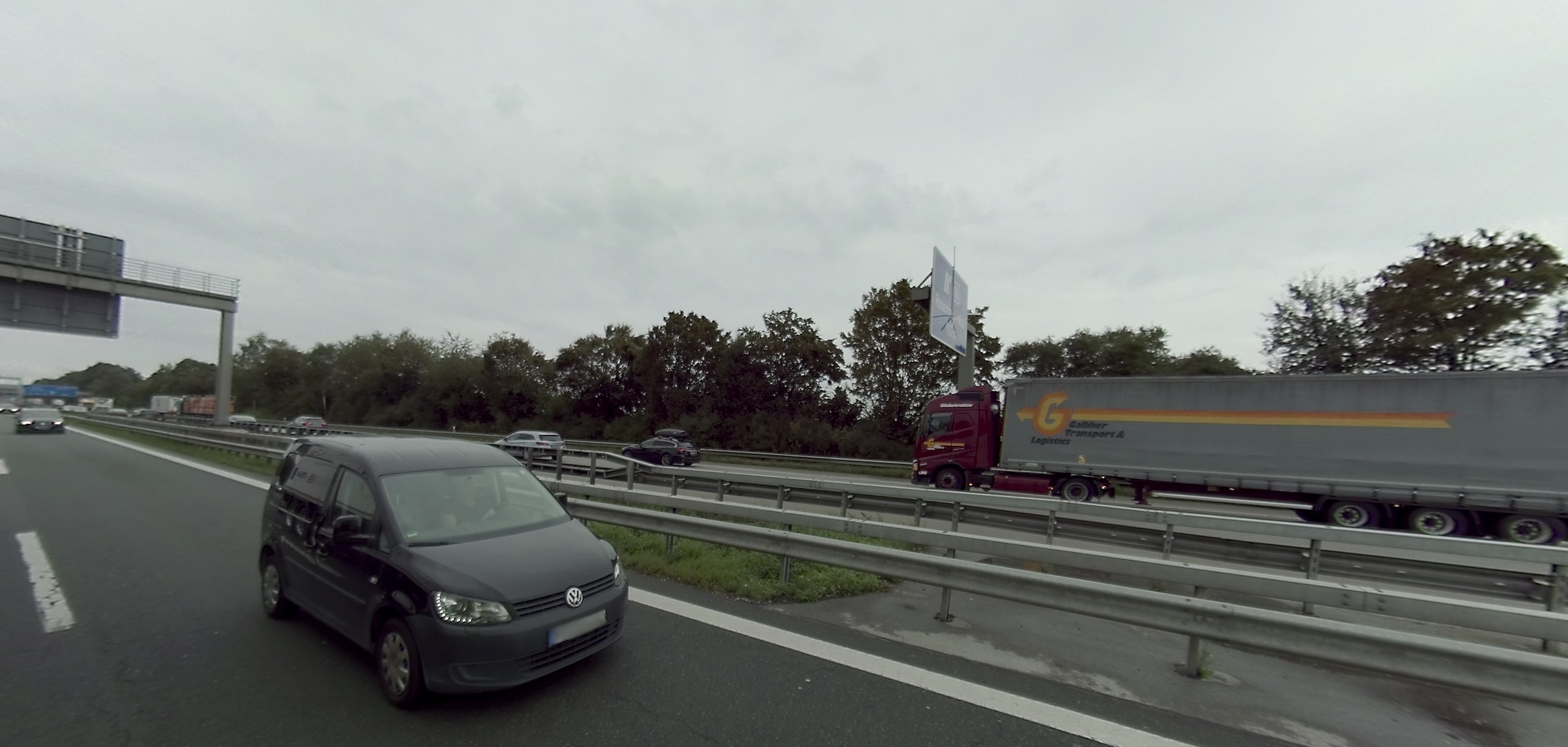}&
                    \includegraphics[width=0.3\textwidth]{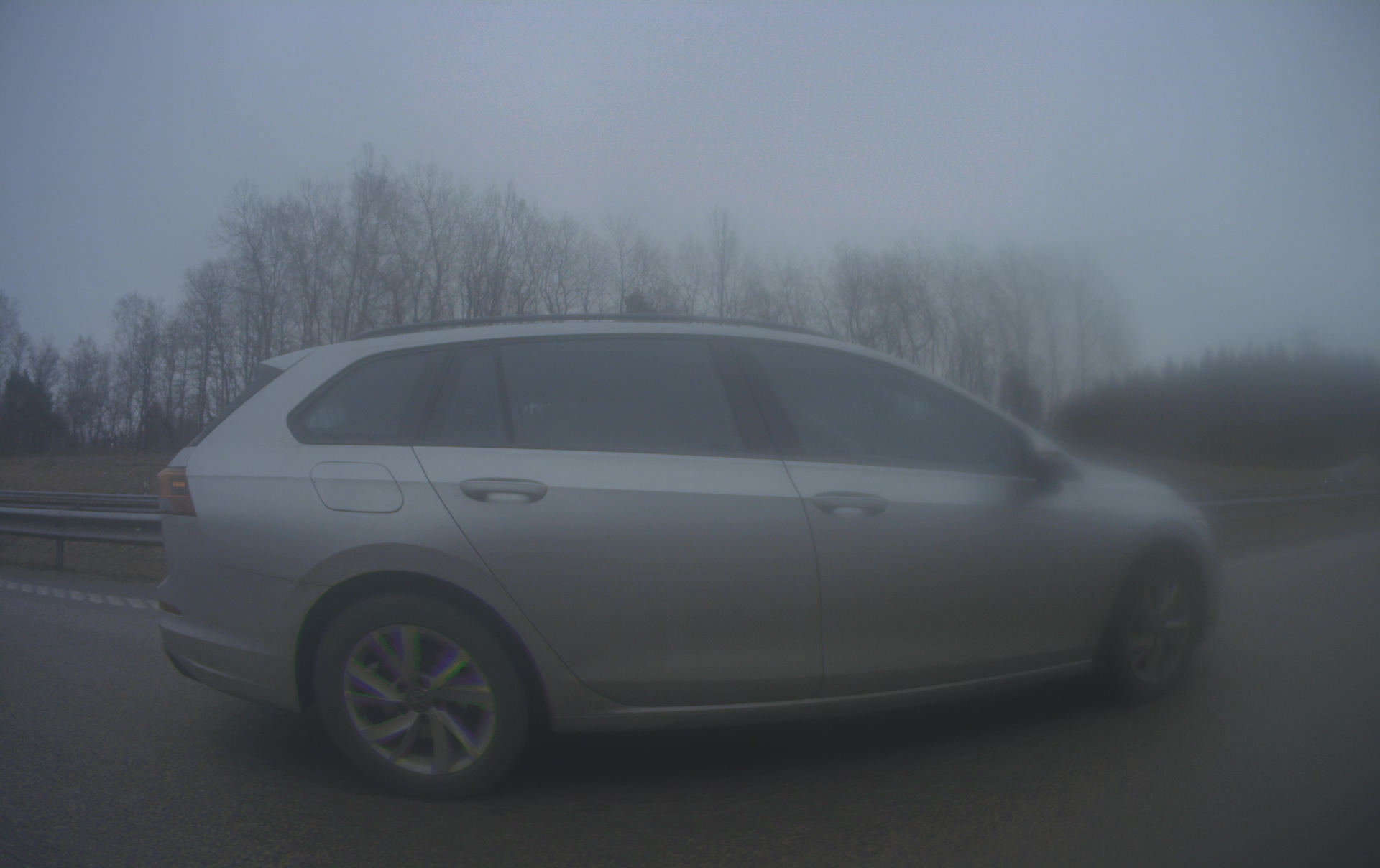}
                \end{tabular}
            \end{subfigure}\\
        
            % ORIGINAL BOXES AND EGO MOTION COMPENSATED POINTS - 3 DATASETS AND 2 VIEWS EACH
            \begin{subfigure}[c]{0.9\textwidth}
                \begin{tabular}{p{0.3\textwidth}p{0.3\textwidth}p{0.3\textwidth}}
                    \includegraphics{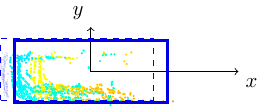}&
                    \includegraphics{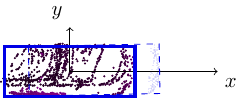}&
                    \includegraphics{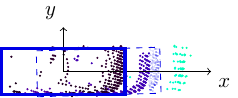}
                    \hspace{0pt}\vspace{5pt}\\
                    \includegraphics{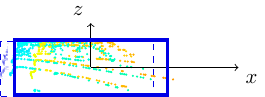}&
                    \includegraphics{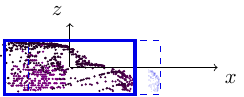}&
                    \includegraphics{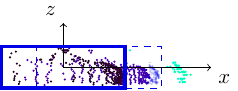}
                    \hspace{0pt}\vspace{5pt}\\
                \end{tabular}
            \end{subfigure}\\

            % CAPTIONS - 3 DATASETS
            \begin{subfigure}[c]{0.9\textwidth}
                \begin{tabular}{p{0.3\textwidth}p{0.3\textwidth}p{0.3\textwidth}}
                    \begin{subfigure}[c]{0.275\textwidth}
                        \caption{Argoverse 2.}
                        \label{fig:argoverse2_ego_mc}
                    \end{subfigure}&
                    \begin{subfigure}[c]{0.275\textwidth}
                        \caption{MAN TruckScenes.}
                        \label{fig:man_truckscenes_ego_mc}
                    \end{subfigure}&
                    \begin{subfigure}[c]{0.275\textwidth}
                        \caption{Our proprietary dataset.}
                        \label{fig:proprietary_ego_mc}
                    \end{subfigure}
                \end{tabular}
            \end{subfigure}
            
        \end{tabular}&

        % COLOR BAR
        \begin{subfigure}[c]{0.1\textwidth}
            \includegraphics{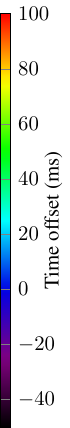}
        \end{subfigure}
        
    \end{tabular}
    \caption{Examples of 3D annotation errors in various datasets. The original \gls{lidar} points are colored by capture time offset relative to the reference time at which the boxes were annotated (see color bar at the right of the figure). The translucent, blue points show where these \gls{lidar} detections would have been at the annotation reference time. Our proposed annotations (dashed-line boxes) are aligned with points with zero time offset, while the original annotations (solid-line boxes) are aligned with point clouds with large time offsets. The position offset between them is an indication of the annotation error that we aim to correct.}
    \label{fig:introduction}
\end{figure*}

For object detection, typical datasets \cite{geiger_kitti_2012,caesar_nuscenes_2020,sun_waymo_2020,wilson_argoverse_2021,alibeigi_zenseact_2023,fent_man_2024} provide bounding box positions at a specific reference timestamp within a \qty{100}{\milli\second} interval, \ie annotation sample, which is a typical scanning interval for a rotating \gls{lidar} sensor. Annotation is typically done by fitting a 3D bounding box to the \gls{lidar} data associated with the edges of the object. However, a rotating \gls{lidar} may detect an object any time within the mentioned \qty{100}{\milli\second} interval. Specifically, during \qty{100}{\milli\second}, objects can move significantly in highly dynamic scenarios, \eg \qty{3}{\meter} for a typical highway speed of \qty{30}{\meter\per\second}. Hence, in general, the ground truth pose of the object differs between the actual time when the \gls{lidar} detected the object and the common reference time at which all boxes are annotated. This discrepancy may lead to large box annotation errors. These are more obviously visible in multi-\gls{lidar} datasets (see \cref{fig:introduction}), but are present in single-\gls{lidar} datasets as well \cite{video_lidar_distortion}.

For example, \cref{fig:argoverse2_ego_mc} shows a 3D box annotation in the Argoverse 2 \cite{wilson_argoverse_2021} dataset. There, the object is moving towards the right of the image, along its $x$-axis. However, the box is aligned with the \gls{lidar} detections at \qty{30}{\milli\second} offset \wrt the box annotation timestamp, hence significantly more to the right than where the object really was at the box annotation timestamp. In turn, \cref{fig:man_truckscenes_ego_mc} shows a similar situation in the MAN TruckScenes dataset. The position of the box corresponds to the location where the object was at about \qty{-35}{\milli\second} offset, hence significantly more to the left than where the object really was at the annotation timestamp. Finally, \cref{fig:proprietary_ego_mc} shows an equivalent setting in our proprietary dataset. The box is aligned with \gls{lidar} detections offset \qty{-20}{\milli\second}, hence significantly more to the left than where the object was at the annotation timestamp. In all these cases, the boxes should have been placed at the location where hypothetical \gls{lidar} detections offset with \qty{0}{\milli\second} \wrt the annotation timestamp (the translucent blue points in \cref{fig:introduction}) would have hit the moving objects, \ie where the dashed-line boxes are drawn.

From inspecting these and more annotations in the datasets, we hypothesize that the original annotations were made by arbitrarily choosing one reference set of sensor detections to fit the boxes to, without taking into consideration neither the time at which those points were captured nor the dynamics of the objects.

An accurate annotation would either place the boxes where the objects were at the annotation timestamp or should have indicated the exact timestamp when each object's position was annotated. We focus on the former, as annotating multiple boxes at a common, pre-defined sample timestamp has been the standard in the autonomous driving field. However, the latter, \ie annotating each individual box at the exact location and timestamp that matches the sensor data, has the potential to be more accurate and lead to better performance in downstream tasks, as the detectors would only need to find a box around existing sensor data, instead of implicitly estimating motion and compensating the sensor data to a predefined timestamp.

In the case we focus on, the position of the box shall be extrapolated to the sample annotation timestamp using the object's dynamics. Since the object dynamics are not known apriori, such mechanism requires estimating several states of the object dynamics, including speed, yaw rate, and acceleration; by using several frames from the sequence. As exemplified above, this estimation may imply a large box displacement, and in general it may come across as if the original detections were not fitting the extrapolated box.

The novelty of our work is three-fold. First, we introduce a principled, motion-model-based annotation correction framework that accounts for actors' motion, enabling physically plausible 3D box placement, a capability absent in current datasets and prior methods. Second, we define, for the first time, quantitative metrics tailored to assess the quality of dynamic object annotations, allowing for systematic diagnosis of annotation errors and benchmarking of correction techniques. Finally, we demonstrate the significant sensitivity of commonly accepted benchmarks to the annotation errors studied. These contributions not only aim to improve annotation fidelity, but also lay the foundation for future studies on the impact of annotation errors on autonomous driving pipelines.

%% file: 2_related_work.tex
\section{Related work} \label{sec:related_work}

The problem of distorted 3D box annotations was first highlighted by \cite{khoche_addressing_2024}. The authors pointed out that the annotations on their proprietary dataset suffered from distortion issues, more so in highly dynamic environments. They provided a method that improved the longitudinal placement of the boxes by estimating their speed. However, this method is limited in that it assumes rectilinear trajectories, it does not enforce consistency with sensor data, and quantitative evaluations are absent.

More recently, \cite{zhang_himo_2025} pointed out that the multiple \gls{lidar} distortion is visible also in popular public datasets, and provided a scene flow method to compensate \gls{lidar} points for target motion so that they become coherent with each other. However, that work focused on the point clouds only; and did not analyze the box annotations, nor did it attempt to improve them. Their metrics, Chamfer Distance Error and Mean Point Error, are applicable to point matching but not to correction of box annotations.

Therefore, to the best of the authors' knowledge, there have been no attempts to holistically correct the 3D bounding box annotations from any automotive dataset nor proposed metrics for this problem.

\hfill\break\noindent{\bf Motion models.} In the object tracking domain \cite{luo2021multiple} there exist multiple motion models that allow estimation of object dynamics such as speed, yaw rate, and acceleration \cite{rong_surveymotionmodels_2003}. Among these, those that can be utilized in a two-dimensional curvilinear trajectory are of particular interest to the autonomous driving domain \cite{schubert_comparisonmotionmodels_2008}; according to which the best-performing motion model for such applications is the \gls{ctra} model \cite{svensson_ctra_2019}. For complex vehicles, such as vehicles with trailers or articulated vehicles, dedicated motion models can also be found in the literature \cite{chen1997modelingarticulated}.

\hfill\break\noindent{\bf Optimization methods.} A suitable way to estimate the states of the box annotations is by means of an optimization formulation. Due to the complexity of the annotation correction problem, which may include continuous and discrete variables as well as non-differentiable functions, a number of global metaheuristic optimization algorithms \cite{talbi2009metaheuristics} shall be considered. All of them can deal with non-differentiable functions (derivative free) and have mechanisms to escape local optima.

The most commonly used such algorithms are as follows. \gls{ps} \cite{hooke1961direct} requires the search boundaries to be defined. It starts from an initial guess and iteratively steps towards a better solution, shrinking the step size if no better solution is reachable. Genetic Evolution \cite{davis_geneticalgorithms_1991} randomly initializes a population of candidate solutions and iteratively evolves them using mechanisms such as selection, crossover, and mutation. Finally, Particle Swarm \cite{eberhart_particleswarm_1995} randomly initializes a population of candidate solutions and iteratively improves them based on their momentum and the currently best known solution locally and globally.

Among these algorithms, \gls{ps} \cite{hooke1961direct} is the most suitable one when a reliable initial guess can be provided, and has been shown to slightly outperform the other mentioned gradient-free methods in some applications \cite{MANSOURI_coverage_2018}.

%% file: 3_methodology.tex
\section{Methodology} \label{sec:methodology}

This section describes the methodology used to solve the problem. \Cref{sec:problem_formulation} formulates the problem to solve. \Cref{sec:cost_ctra,sec:cost_lidar,sec:cost_dist} define the three terms of the tailored objective function, respectively. \Cref{sec:cost_total} aggregates these terms into one single objective function. Finally, \cref{sec:metrics} defines the evaluation metrics for the problem.

\subsection{Problem formulation} \label{sec:problem_formulation}

The goal of this work is to correct the pose of existing box annotations, so that they better match the collection of sensor (typically, \gls{lidar}) detections associated with the object over time and compensated for object dynamics, while maintaining a physically feasible trajectory of the boxes.

Although size attributes might also carry errors, they are expected to be orders of magnitude lower and less relevant. Furthermore, their annotation is not always possible to do based on sensor data; \eg when only a fraction of an object is observed, and thus some dimensions are unobservable; some expected standard sizes are typically annotated instead. Therefore, we focus on the errors that are caused purely by mishandling of temporal sensor data, namely those in the poses, and leave the size attributes as is.

The inputs to the problem are three: the original box annotations, the sensor 3D point clouds, and the ego vehicle poses. First, let the annotated bounding boxes for every track ID $\alpha$ (constant along the whole sequence for each given object) and annotation sample $i$,
\begin{equation}
    \bm{B}_{\alpha,i} = 
    \begin{bmatrix}
        x_{\alpha,i} & y_{\alpha,i} & z_{\alpha,i} & \theta_{\alpha,i} & L_{\alpha,i} & W_{\alpha,i} & H_{\alpha,i}
    \end{bmatrix}^T ,
\end{equation}
where $x$, $y$ and $z$ refer to the 3D coordinates of the centroid of the box, $\theta$ refers to the orientation of the box about the $z$-axis (i.e., yaw angle) and $L$, $W$ and $H$ refer to length, width, and height, respectively.

Second, let the sensor data, expressed in the global frame of reference to prevent distortion due to ego motion,
\begin{equation}
    \bm{P}_{i,k} =
    \begin{bmatrix}
        p^x_{i,k} & p^y_{i,k} & p^z_{i,k} & \Delta t_{i,k}
    \end{bmatrix}^T ,
\end{equation}
for every annotation sample $i$ and detection $k$, where $p^x$, $p^y$, and $p^z$ refer to the 3D coordinates of a point, and $\Delta t$ refers to the time difference between the time at which a point was captured and the annotation reference timestamp.

Finally, let the ego vehicle's 2D pose,
\begin{equation}
    \bm{E}_i =
    \begin{bmatrix}
        E_i^x & E_i^y & E_i^\theta
    \end{bmatrix}^T,
\end{equation}
for every annotation sample $i$.

For the sake of simplicity, we only formulate the pose correction problem in a 2D, bird's-eye view representation, which is where most of the annotation error is expected for perfectly flat driving surfaces. It would be needed to extend the problem formulation to 3D when handling, for instance, data from hilly terrain.

Thus, the goal is to estimate every box's 2D pose, speed, yaw rate, and acceleration at their corresponding annotation reference timestamp; while the bounding box's $z$-coordinate, as well as the dimensions $L$, $W$ and $H$ are assumed to remain constant. More specifically, we seek to estimate for every track ID $\alpha$ and sample $i$ the state vector
\begin{equation} \label{eq:state_vector_to_be_estimated}
    \bm{X}_{\alpha,i} = 
    \begin{bmatrix}
        x_{\alpha,i} & y_{\alpha,i} & \theta_{\alpha,i} & s_{\alpha,i} & \omega_{\alpha,i} & a_{\alpha,i}
    \end{bmatrix}^T ,
\end{equation}
where the variables $s$ and $a$ refer to linear speed and acceleration, respectively, along the box's $x$-axis; and $\omega$ refers to the box's yaw rate; such that an objective function, $L$, is minimized independently for each track ID $\alpha$, along all annotation samples $i$ of a sequence:
\begin{equation} \label{eq:loss_function}
    \bm{\hat{X}}_\alpha = \argmin_{\bm{X}_\alpha} L \left( \bm{B}_\alpha, \bm{P}, \bm{E} \right) .
\end{equation}
The objective function at \cref{eq:loss_function} is defined in \cref{sec:cost_ctra,sec:cost_lidar,sec:cost_dist,sec:cost_total}.

\subsection{Motion model} \label{sec:cost_ctra}

As the annotated objects correspond to real objects, often vehicles, that move; a motion model shall be enforced in the optimization formulation. Since these objects may move rectilinearly, curvilinearly, or a combination of both, with varying speed profiles; the motion model must account for linear speed, acceleration, and angular speed. To this end, a \gls{ctra} \cite{svensson_ctra_2019} motion model was selected.

The \gls{ctra} equations \cite{svensson_ctra_2019} define state transition over a time interval $\Delta t = t - t_0$ as follows:
\begin{subequations}
\begin{align}
\begin{split} \label{eq:ctra_dx}
    \Delta x =& \frac{1}{\omega_0^2} \left(s_t \omega_0 \sin\left(\theta_t\right) + a_0 \cos\left(\theta_t\right)\right. \\
          & - s_0 \omega_0 \sin\left(\theta_0\right) - a_0 \cos\left(\theta_0\right) \left.\right) ,
\end{split} \\
\begin{split} \label{eq:ctra_dy}
\Delta y =& \frac{1}{\omega_0^2} \left(-s_t \omega_0 \cos\left(\theta_t\right) + a_0 \sin\left(\theta_t\right)\right. \\
          & + s_0 \omega_0 \cos\left(\theta_0\right) - a_0 \sin\left(\theta_0\right) \left.\right) ,
\end{split} \\
\label{eq:ctra_dth}
\Delta \theta =& \omega_0 \cdot \Delta t ,
\\
\label{eq:ctra_ds}
\Delta s =& a_0 \cdot \Delta t ,
\\
\label{eq:ctra_dw}
\Delta \omega =& 0 ,
\\
\label{eq:ctra_da}
\Delta a =& 0 .
\end{align}
\end{subequations}
For $\omega_0$ close or equal to zero, a numerically stable approximation of \cref{eq:ctra_dx,eq:ctra_dy} can be obtained as described in \cite{svensson_ctra_2019}.

The \gls{ctra} equations can be rewritten in vector form:
\begin{equation}
     \bm{\Delta Y}_0^t =
     \begin{bmatrix}
        \Delta x & \Delta y & \Delta \theta & \Delta s & \Delta \omega & \Delta a
    \end{bmatrix}^T .
\end{equation}
The \gls{ctra} constraint aims to minimize the difference between the predicted states and the optimization variable states. More specifically, from each state node $i$, a \gls{ctra} prediction is made to the time of the subsequent node $i+1$,
\begin{equation}
    \bm{Y}_i^{i+1} = \bm{X}_i + \bm{\Delta Y}_i^{i+1} .
\end{equation}
This prediction is compared to the optimization variable at node $i+1$ using a weighted squared error,
\begin{equation}
    \epsilon_{i+1}^m = \left(\bm{X}_{i+1} - \bm{Y}_i^{i+1}\right) ^ T \bm{\beta_m} \left(\bm{X}_{i+1} - \bm{Y}_i^{i+1}\right) ,
    \label{eq:errorCTRA}
\end{equation}
where $\bm{\beta_m}$ is the inverse covariance matrix for the \gls{ctra} motion model, specified in \cref{sec:implementation}. This term adds a penalty given by how much the movement differs from the ideal \gls{ctra} movement. Note that this formulation requires that tracks consist of at least two boxes along a sequence.

\subsection{Sensor detections} \label{sec:cost_lidar}

Adding constraints on the boxes being consistent with the sensor detections is key to ensuring that the annotation errors are corrected. The following describes the components needed for it.

\hfill\break\noindent{\bf Point-to-box association.} Because, in general, datasets do not provide labels for each sensor detection, the ones that fall within certain limits of each box are considered susceptible to belonging to the object. Thus, each original box is inflated so as to include all the detections that can reasonably belong to the object, but no detections from other objects. This depends on the typically expected distance between objects, which might be larger at high vehicle speeds than in a slow urban traffic.

Thus, we inflate each box on both ends along its $x$-axis (rear and front) proportionally to its speed,
$\Delta L_{\alpha,i}^\text{rear} = \Delta L_{\alpha,i}^\text{front} = \qty{1}{\meter} + \Delta T_\text{sensor}\cdot s_{\alpha,i}$,
where $\Delta T_\text{sensor}$ is the scanning period of the sensor. Moreover, the width is inflated a constant amount on each side (left and right),
$\Delta W^\text{left} = \Delta W^\text{right} = \qty{0.5}{\meter}$,
and the height is deflated a constant amount at the bottom to prevent interference from ground detections in the optimization,
$\Delta H^\text{bottom} = \qty{-0.2}{\meter}$,
while the height remains untouched at the top. This results in a set of sensor detections that are associated with each box, $\bm{P}_{i,k}^\alpha$.

\hfill\break\noindent{\bf Target motion compensation.} Since, in general, annotated targets move over the course of a sample, causing point cloud distortion, these point clouds need to be compensated for target motion.

The points that have been associated with a box are compensated for target motion using the estimated dynamics of the box (\cref{eq:state_vector_to_be_estimated}) at each time step, using the \gls{ctra} formulas from \cref{eq:ctra_dx,eq:ctra_dy} while setting $\Delta t$ to each point's time offset \wrt the annotation timestamp. This is,
\begin{subequations}
\begin{align}
\begin{split} \label{eq:target_mc_x}
    \tilde{p}^x_{i,k} =& p^x_{i,k} + \frac{1}{\omega_{\alpha,i}^2} \left(s_t \omega_{\alpha,i} \sin\left(\theta_t\right) + a_{\alpha,i} \cos\left(\theta_t\right)\right. \\
          & - s_{\alpha,i} \omega_{\alpha,i} \sin\left(\theta_{\alpha,i}\right) - a_{\alpha,i} \cos\left(\theta_{\alpha,i}\right) \left.\right) ,
\end{split} \\
\begin{split} \label{eq:target_mc_y}
    \tilde{p}^y_{i,k} =& p^y_{i,k} + \frac{1}{\omega_{\alpha,i}^2} \left(-s_t \omega_{\alpha,i} \cos\left(\theta_t\right) + a_{\alpha,i} \sin\left(\theta_t\right)\right. \\
          & + s_{\alpha,i} \omega_{\alpha,i} \cos\left(\theta_{\alpha,i}\right) - a_{\alpha,i} \sin\left(\theta_{\alpha,i}\right) \left.\right) ;
\end{split}
\end{align}
\end{subequations}
where
\begin{subequations}
\begin{align}
\begin{split} \label{eq:target_mc_x_aux}
    \theta_t =& \theta_{\alpha,i} + \omega_{\alpha,i} \cdot \Delta t_{i,k} ,
\end{split} \\
\begin{split} \label{eq:target_mc_y_aux}
    s_t =& s_{\alpha,i} + a_{\alpha,i} \cdot \Delta t_{i,k} .
\end{split}
\end{align}
\end{subequations}
A similar set of equations can be derived for the approximation for low yaw rates previously mentioned in this section.

This results in a motion compensated point cloud at the annotation reference timestamp,
\begin{equation}
    \bm{\tilde{P}}_{i,k}^\alpha =
    \begin{bmatrix}
        \tilde{p}^x_{i,k} & \tilde{p}^y_{i,k} & p^z_{i,k} & \Delta t_{i,k}
    \end{bmatrix}^T ,
\end{equation}
for every sample $i$ and point within the sample $k$.

\hfill\break\noindent{\bf Inlier ratio.} The optimal box annotation will be such that the ratio of associated, motion compensated points within itself is maximal. To this end, we define the cost
\begin{equation}
    \epsilon_{\alpha,i}^\text{inlier} = 1 - \frac{\bigm|\bm{\tilde{P}}_{i,k}^\alpha \in \bm{X}_{\alpha,i}\bigm|}{\bigm|\bm{\tilde{P}}_{i,k}^\alpha\bigm|} .
    \label{eq:error_points_ratio}
\end{equation}

\hfill\break\noindent{\bf Fitness.} Maximizing the inlier ratio alone is not enough to guarantee optimal box placement. In addition, the boxes must fit the points as tightly as possible. Thus, we define a cost term that rewards the points being as close as possible to any face of the box.

Let the two orthogonal vectors that define the box in the 2D, bird's-eye view space, $\bm{c}_x$ and $\bm{c}_y$. Project each point $\bm{\tilde{P}}_{i,k}^\alpha$ to both orthogonal vectors. By basic algebra, when these projections are within the interval $\left[0, 1\right]$, the point lies within the box. More importantly, the closer any of these projections is to 0 or 1, the closer the point is to a face of the box. Because it is not known beforehand which face of the box the points should fit, only the best-fitting face of the four will be evaluated. Thus, the cost term is such that the average of fitness between the associated points and the box is maximized,
\begin{equation}
    \begin{split}
        \epsilon_{\alpha,i}^\text{fit} = \frac{1}{\bigm|\bm{\tilde{P}}_{i,k}^\alpha\bigm|} \sum_{\bm{\tilde{P}}_{i,k}^\alpha} \left(2 \cdot \min \left(\bm{\tilde{P}}_{i,k}^\alpha \cdot \bm{c}_x, \bm{\tilde{P}}_{i,k}^\alpha \cdot \bm{c}_y,\right.\right.\\
        1 - \bm{\tilde{P}}_{i,k}^\alpha \cdot \bm{c}_x,
        \left.\left. 1 - \bm{\tilde{P}}_{i,k}^\alpha \cdot \bm{c}_y\right)\right)^2 .
    \end{split}
    \label{eq:error_points_fit}
\end{equation}
A set of points that lie on either face of the box will result in $\epsilon_{\alpha,i}^\text{fit} = 0$; whereas a set of points that lies in the center of the box will result in $\epsilon_{\alpha,i}^\text{fit} = 1$, which is the local maximum within the box; and a point that lies outside of the box will result in $\epsilon_{\alpha,i}^\text{fit} > 0$ and will quadratically increase boundlessly with distance away from the box edges.

\hfill\break\noindent{\bf Overall sensor detections loss.} Overall, the sensor detections cost becomes the sum of \cref{eq:error_points_ratio,eq:error_points_fit}, weighted by the constant $\beta_s$, defined in \cref{sec:implementation}. This reads
\begin{equation}
    \epsilon_{\alpha,i}^s = \beta_s \left( \epsilon_{\alpha,i}^\text{inlier} + \epsilon_{\alpha,i}^\text{fit} \right) .
    \label{eq:errorLIDAR}
\end{equation}

\subsection{Distance to ego} \label{sec:cost_dist}

To make the box placement more accurate, we add a term that penalizes the box from being close to the ego vehicle. The rationale is that the sensors, which are mounted on the ego vehicle, will detect the closest faces of the objects, and not the farthest ones. Thus, this term prevents the previous terms from fitting a wrong face of the boxes to the points, especially in cases in which only one face of the box was sensed. The term reads
\begin{equation} \label{eq:errorDISTEGO}
    \epsilon_{\alpha,i}^d = - \beta_d \cdot \lVert\begin{bmatrix}
        x_{\alpha,i} & y_{\alpha,i}\end{bmatrix} - \begin{bmatrix}
        E^x_i & E^y_i\end{bmatrix}\rVert ,
\end{equation}
where $\beta_d$ is a constant weight that should make this term less influential than the other terms in the optimization, and will be defined in \cref{sec:implementation}.

\subsection{Objective function} \label{sec:cost_total}

The overall objective function for a track $\alpha$ becomes the aggregation of \cref{eq:errorCTRA,eq:errorLIDAR,eq:errorDISTEGO} along its trajectory,
\begin{equation}
    \bm{\hat{X}}_\alpha = \argmin_{\bm{X}_\alpha}
    \sum_{\bm{B}_{\alpha,i}\setminus\bm{B}_{\alpha,1}} \epsilon_{\alpha,i}^m + \sum_{\bm{B}_{\alpha,i}} \left( \epsilon_{\alpha,i}^s + \epsilon_{\alpha,i}^d \right) .
    \label{eq:total_cost_function}
\end{equation}
This objective function can be applied independently to different trajectories, even from different sequences.

\begin{figure}[htbp]
\centering
\includegraphics{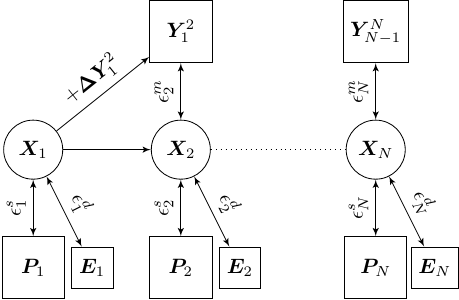}
\caption{A diagram that illustrates the full optimization problem. The states to estimate, $X_i$, are within circles, whereas the variables that are used to compute the error are within squares. The \gls{ctra} motion model, $Y_i^{i+1}$, aims to enforce a physically plausible trajectory by minimizing the difference between the predicted states from the $i$th to the $(i+1)$th time step and the optimization variable at the $(i+1)$th time step. In turn, the point cloud, $P_i$, and distance to ego, $E_i$, error terms are applied at every time step and aim to place the corrected boxes coherently with the sensor data.}
\label{fig:optimization_diagram}
\end{figure}

The complete optimization problem is illustrated in \cref{fig:optimization_diagram} for a track composed of $N$ annotated boxes along its trajectory.

\subsection{Metrics} \label{sec:metrics}

Since no metrics have been defined for the annotation correction problem until the time of writing (see \cref{sec:related_work}), we propose the following as a basis for a quantitative evaluation of both the performance of our method and 3D box annotation quality.

\hfill\break\noindent{\bf Inlier points difference.} The \gls{ipd} proposed metric calculates the relative difference in the absolute number of sensor detections compensated for motion that lie within the original and corrected boxes,
\begin{equation}
    \text{IPD} = \frac{\bm{\tilde{P}} \in \bm{\hat{X}} - \bm{\tilde{P}} \in \bm{B}}{\bm{\tilde{P}} \in \bm{B}} .
\end{equation}
Naturally, a functioning annotation correction method is expected to produce a positive \gls{ipd}, which means that the corrected boxes capture more compensated points than the original boxes and thus are more appropriately positioned.

\hfill\break\noindent{\bf Error distribution and correlation.} Once the corrected boxes are estimated, the difference between them and the original boxes can be calculated and regarded as a measure of the error of the original boxes. Our proposed metrics compute the \gls{ede} and its probability distribution; as well as the \gls{dede} along the box's $x$ and $y$ axes, to determine if the error of the original boxes is correlated with their traveling direction, which would reveal systematic errors due to incorrect handling of object dynamics. We propose a quantification of the latter, namely \gls{sdede}, calculated as 3 times the standard deviation \cite{replaces2004protocols} of the respective distributions. Additionally, we propose computing the \gls{ede} against a set of speed and distance to ego intervals to discover eventual correlations in the datasets between error, on the one hand, and speed and distance, respectively, on the other hand.

%% file: 4_experiments.tex
\section{Experiments} \label{sec:experiments}

Our experiments are carried out on 10 sequences of each of Argoverse 2, MAN TruckScenes, and our proprietary datasets. The 10 chosen sequences of Argoverse 2 are the ones with the highest average object speed, calculated by naively dividing box distances by time differences, and can be found in our code; while the 10 sequences of MAN TruckScenes correspond to its mini dataset version. We limit the number of sequences because of the intractable amount of computation time that would be needed to correct complete datasets with our method, and because the main goal of our work is to raise and quantify an underexplored problem rather than solving it completely or efficiently.

The reason why we choose these datasets is that they provide a sufficient amount of information, including explicit time offset per individual \gls{lidar} detection and 3D box annotations along multiple frames within the sequences. Most other datasets mentioned in \cref{sec:introduction} do not provide this required information and thus make the systematic annotation errors that we tackle very difficult to even diagnose at all.

Also note that all the datasets to which our method is applied in this work happen to be multi-\gls{lidar}. However, our method is general enough to handle single or multiple \gls{lidar} and any other active sensor datasets.

\subsection{Implementation} \label{sec:implementation}

As discussed in \cref{sec:related_work}, the \gls{ps} \cite{hooke1961direct} algorithm is the most reasonable choice for our experiments, since our problem setting allows us to provide a reasonably good initial guess from the original annotations. This algorithm is gradient-free and therefore we use it to accommodate the objective function of \cref{eq:total_cost_function}. Any other gradient-free optimization algorithms (see \cref{sec:related_work}) can accommodate the same objective function and are compatible for solving this problem as well. Our method was developed in Python relying on the \gls{ps} implementation from the pymoo \cite{pymoo} library.

The states $x$, $y$, and $\theta$ of the original box annotations were used as the initial value for the corresponding optimization variables, while the initial value for speed was established by naively dividing the distance by the time difference in the original box annotations, and the initial values for yaw rate and acceleration were set at zero.

Additionally, the search space was bounded from the initial values with \qty{\pm 5}{\meter} for $x$ and $y$; \qty[parse-numbers=false]{\pm\nicefrac{\pi}{16}}{\radian} for $\theta$; \qty{\pm 40}{\meter\per\second} for speed; \qty[parse-numbers=false]{\pm\nicefrac{\pi}{8}}{\radian\per\second} for yaw rate; and \qty{\pm 20}{\meter\per\second\squared} for acceleration.

Finally, the optimization weights were set as follows:
$\bm{\beta_m} = 10 \cdot f_a \cdot \bm{\mathbb{I}_6}$ with $f_a$ being the box annotation frequency (\qty{10}{\hertz} for Argoverse 2 and our proprietary dataset and \qty{2}{\hertz} for MAN TruckScenes),
$\beta_s = 10^{3}$, and
$\beta_d = \qty[parse-numbers=false]{10^{-4}}{\per\meter}$;
so the ratio and placement of points within the boxes are the most influential, followed by the \gls{ctra} motion model, followed by the distance of boxes to ego.

\subsection{Qualitative results}

In the first experiment, we test our method in the samples that correspond to \cref{fig:introduction}. The resulting corrected boxes can be seen in \cref{fig:introduction} as dashed-line boxes along with the points compensated for the track's motion as translucent blue points. The corrected boxes are much closer to their ideal position than the original annotated box, as they fit the auxiliary motion-compensated sensor detections tightly.

\subsection{Quantitative results}

\Cref{tab:experiments_combined_datasets} shows results for the mentioned datasets and sequences. All cases show a clear improvement in the \gls{ipd}.

\begin{table}[htbp]
    \centering
    \begin{tabular}{|c|r|r|}
        \hline
        \textbf{Dataset} &
        \multicolumn{1}{c|}{\textbf{\makecell{Average\\Speed}}} &
        \multicolumn{1}{c|}{\textbf{\makecell{IPD}}} \\
        \hline
        Argoverse 2 &
        \qty[per-mode = symbol]{11.12}{\meter\per\second} &
        +17.53 \% \\
        \hline
        MAN TruckScenes &
        \qty[per-mode = symbol]{13.58}{\meter\per\second} &
        +30.45 \% \\
        \hline
        Proprietary &
        \qty[per-mode = symbol]{15.61}{\meter\per\second} &
        +27.99 \% \\
        \hline
    \end{tabular}
    \caption{Average Speed and \gls{ipd} on 10 sequences of each of 3 datasets. Only boxes whose speed is at least \qty[per-mode = symbol]{3}{\meter\per\second} are included.}
    \label{tab:experiments_combined_datasets}
\end{table}

\begin{figure*}[htbp]
    \centering
    \begin{tabular}{cc}
        \begin{subfigure}[b]{0.45\textwidth}
            \centering
            \includegraphics{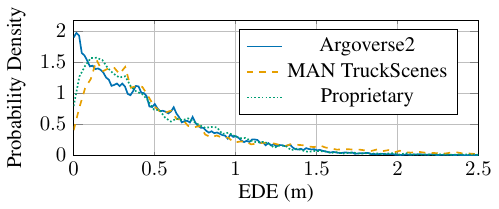}
            \caption{Distribution of \gls{ede} on the 3 studied datasets.}
            \label{fig:probability_density_euclidean_distance_error}
        \end{subfigure}&
        \begin{subfigure}[b]{0.45\textwidth}
            \centering
            \includegraphics{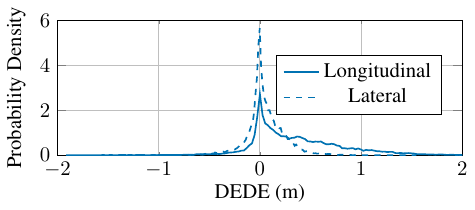}
            \caption{Distributions of \gls{dede} on the Argoverse 2 dataset.}
            \label{fig:probability_density_euclidean_distance_error_decoupled_argoverse2}
        \end{subfigure}\\
        \begin{subfigure}[b]{0.45\textwidth}
            \centering
            \includegraphics{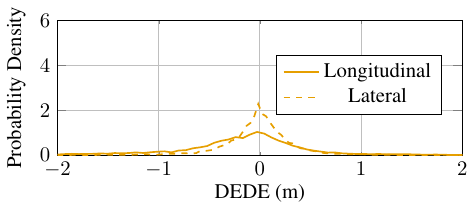}
            \caption{Distributions of \gls{dede} on the MAN TruckScenes dataset.}
            \label{fig:probability_density_euclidean_distance_error_decoupled_man_truckscenes}
        \end{subfigure}&
        \begin{subfigure}[b]{0.45\textwidth}
            \centering
            \includegraphics{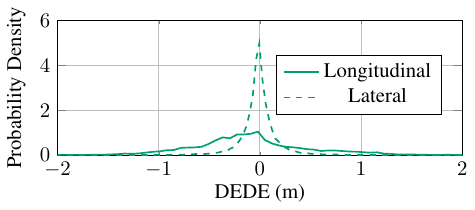}
            \caption{Distributions of \gls{dede} on our proprietary dataset.}
            \label{fig:probability_density_euclidean_distance_error_decoupled_proprietary}
        \end{subfigure}
    \end{tabular}
    \caption{Distributions of error of the original box annotations \wrt the corrected boxes in the selected sequences of the 3 examined datasets. Only boxes whose speed is at least \qty[per-mode = symbol]{3}{\meter\per\second} are included in the plots.}
    \label{fig:distribution_error}
\end{figure*}

\Cref{fig:distribution_error} shows the distribution of \gls{ede} (\cref{fig:probability_density_euclidean_distance_error}) and \gls{dede} along the $x$ and $y$ axes of the boxes (\cref{fig:probability_density_euclidean_distance_error_decoupled_argoverse2,fig:probability_density_euclidean_distance_error_decoupled_man_truckscenes,fig:probability_density_euclidean_distance_error_decoupled_proprietary}, respectively, for each dataset), taken as the position of the original boxes minus the position of the corrected boxes. Moreover, \cref{tab:spread} shows the \gls{sdede} along the $x$ and $y$ axes of the boxes. From \cref{fig:probability_density_euclidean_distance_error}, it can be seen that the boxes in the original datasets are incorrectly placed by up to \qty{2.5}{\meter}. In addition, according to \cref{fig:probability_density_euclidean_distance_error_decoupled_argoverse2,fig:probability_density_euclidean_distance_error_decoupled_man_truckscenes,fig:probability_density_euclidean_distance_error_decoupled_proprietary,tab:spread}, the strongest component of the errors is observed along the $x$ axes of the boxes, which define their direction of travel. This confirms the hypothesis that most of the error is introduced due to incomplete handling of sensor data in dynamic objects during the annotation process.

\begin{table}[htbp]
    \centering
    \begin{tabular}{|c|r|r|r|}
        \hline
        \textbf{Dataset} &
        \multicolumn{1}{c|}{\textbf{\makecell{SDEDE-x}}} &
        \multicolumn{1}{c|}{\textbf{\makecell{SDEDE-y}}} \\
        \hline
        Argoverse 2 &
        \qty[per-mode = symbol]{1.36}{\meter} &
        \qty[per-mode = symbol]{0.55}{\meter} \\
        \hline
        MAN TruckScenes &
        \qty[per-mode = symbol]{2.54}{\meter} &
        \qty[per-mode = symbol]{1.19}{\meter} \\
        \hline
        Proprietary &
        \qty[per-mode = symbol]{2.17}{\meter} &
        \qty[per-mode = symbol]{0.69}{\meter} \\
        \hline
    \end{tabular}
    \caption{\gls{sdede} on 10 sequences of each of 3 datasets. Only boxes whose speed is at least \qty[per-mode = symbol]{3}{\meter\per\second} are included.}
    \label{tab:spread}
\end{table}

Furthermore, \cref{fig:probability_density_euclidean_distance_error_decoupled_argoverse2,fig:probability_density_euclidean_distance_error_decoupled_man_truckscenes,fig:probability_density_euclidean_distance_error_decoupled_proprietary} show that the longitudinal error is skewed towards different directions for different datasets. For Argoverse 2, it is skewed towards positive errors; whereas for MAN TruckScenes and our proprietary dataset, it is skewed towards negative errors. This is explained by the fact that the annotation samples in the Argoverse 2 dataset include sensor detections with time offsets between \qty{0}{\milli\second} and \qty{100}{\milli\second}; whereas the MAN TruckScenes and our proprietary dataset's are between \qty{-50}{\milli\second} and \qty{50}{\milli\second}, but scanning the road section where most traffic participants are (left of ego vehicle) at negative time offsets and outside of the road (right of ego vehicle) at positive time offsets. Whenever a box was fitted to a set of \gls{lidar} points with positive time offsets, it means that it is ahead of where it should have been (as in \cref{fig:argoverse2_ego_mc}), resulting in a positive original minus corrected longitudinal distance error, which can be seen in \cref{fig:probability_density_euclidean_distance_error_decoupled_argoverse2}. In contrast, when a box was fitted to a set of \gls{lidar} points with negative time offsets, it means that it is behind where it should have been (as in \cref{fig:man_truckscenes_ego_mc,fig:proprietary_ego_mc}), resulting in a negative original minus corrected longitudinal distance error, which can be seen in \cref{fig:probability_density_euclidean_distance_error_decoupled_man_truckscenes,fig:probability_density_euclidean_distance_error_decoupled_proprietary}.

\begin{figure*}[htbp]
    \centering
    \begin{subfigure}[b]{0.45\textwidth}
        \centering
        \includegraphics{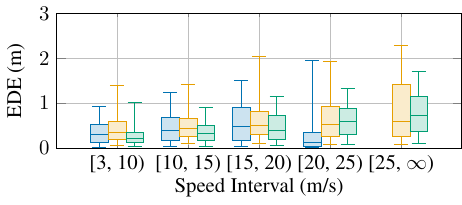}
        \caption{\gls{ede} grouped by speed intervals.}
        \label{fig:error-vs-speed}
    \end{subfigure}
    \begin{subfigure}[b]{0.45\textwidth}
        \centering
        \includegraphics{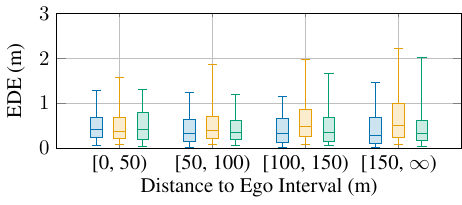}
        \caption{\gls{ede} grouped by distance to ego intervals.}
        \label{fig:error-vs-distance-to-ego}
    \end{subfigure}
    \caption{Box annotation error grouped by intervals. Within each interval, from left to right, the boxes represent the Argoverse 2 (blue), MAN TruckScenes (orange), and our proprietary (green) datasets, respectively. Within each box, from bottom to top, are represented: 5\% percentile, 25\% percentile, median, 75\% percentile, and 95\% percentile. Only objects whose speed is at least \qty[per-mode = symbol]{3}{\meter\per\second} are included.}
    \label{fig:error-vs-speed-and-distance-to-ego}
\end{figure*}

Finally, \cref{fig:error-vs-speed-and-distance-to-ego} shows the relationship between the \gls{ede} and both the speed of the annotated tracks and the distance between the tracks and the ego vehicle, for the 3 studied datasets. Specifically, in \cref{fig:error-vs-speed} it can be observed that the annotation error tends to increase with speed. This, again, confirms the hypothesis that the error is systematic and caused by not considering object dynamics during the annotation process, as it is amplified the more dynamic the objects are. Additionally, \cref{fig:error-vs-distance-to-ego} shows that the annotation error is not correlated with the object distance to ego. More notably, near objects, which are typically more critical to an autonomous driving task than farther objects, are as susceptible to annotation errors as any other object in the scene.

\subsection{Impact on downstream tasks and benchmarks}
While it is beyond the scope of this paper to evaluate or correct the annotation error in complete datasets, we design an experiment to get an indication of the impact that the errors may have on downstream tasks and benchmarks.

Based on the limited sequences studied, the annotation error could be on the order of magnitude indicated in \cref{tab:spread}. In our experiment, we take as an example the Argoverse 2 dataset and the object detection network VoxelNeXt \cite{chen2023voxelnext} with its public checkpoint weights. We assume that the original annotations are error-free, then check if and how much the object detection benchmark metrics \cite{argoverse2metrics} change when the validation annotations are modified by injecting random errors according to the distribution of \cref{tab:spread}.

\begin{table}[htbp]
    \centering
    \begin{tabular}{|c|r|r|r|r|}
        \hline
        \textbf{Method} &
        \textbf{Val. data} &
        \multicolumn{1}{c|}{\textbf{\makecell{AP}}} &
        \multicolumn{1}{c|}{\textbf{\makecell{ATE}}} &
        \multicolumn{1}{c|}{\textbf{\makecell{CDS}}} \\
        \hline
        VoxelNeXt \cite{chen2023voxelnext} &
        Original &
        \qty[per-mode = symbol]{30.5}{} &
        \qty[per-mode = symbol]{0.461}{\meter} &
        \qty[per-mode = symbol]{23.0}{} \\
        \hline
        \makecell{FSD \cite{fan2023super}} &
        Original &
        \qty[per-mode = symbol]{28.2}{} &
        - &
        \qty[per-mode = symbol]{22.7}{} \\
        \hline
        VoxelNeXt \cite{chen2023voxelnext} &
        Modified &
        \qty[per-mode = symbol]{26.0}{} &
        \qty[per-mode = symbol]{0.650}{\meter} &
        \qty[per-mode = symbol]{18.7}{} \\
        \hline
    \end{tabular}
    \caption{The impact that modifying the Argoverse 2 validation split according to the error distribution of \cref{tab:spread} has on metrics and benchmarks is larger than the amount by which VoxelNeXt became state-of-the-art ahead of FSD.}
    \label{tab:impact}
\end{table}

The results can be seen in \cref{tab:impact}: the Argoverse 2 metrics Average Precision (AP), Average Translation Error (ATE), and Composite Detection Score (CDS) are degraded by an amount greater than the margin by which this network became state-of-the-art on the Argoverse 2 object detection benchmark. Likewise, one could arguably expect the performance in real scenarios to be degraded by similar amounts due to the existing errors in the training splits of datasets with which networks are trained.

%% file: 5_conclusion_and_future_work.tex
\section{Conclusion and future work} \label{sec:conclusion}

We have found that 3D box annotations from state-of-the-art automotive datasets carry an error of up to \qty{2.5}{\meter}, caused by incomplete handling of temporal sensor scanning in dynamic scenes. We have proposed an optimization method to correct them so that they better reflect the actual ground truth at the annotation timestamps provided in the datasets. Its applicability extends to any dataset consisting of timestamped 3D sensor detections and box annotations. The results show that such systematic annotation errors occur mainly along the traveling direction of the annotated objects and that it scales with the objects' speed. These errors appear to significantly affect benchmark metrics: their impact goes beyond the typical performance difference among different state-of-the-art methods in the benchmarks.

However, a systematic evaluation of the impact on real-world performance would require correcting complete datasets and retraining and testing models on them, which is left for future work. As future work also remains speeding up the optimization process by using gradient-based optimization, expanding the method to correct potentially wrong box size attributes, and integrating 3D motion into the optimization formulation. Potentially, our method can be integrated into annotation pipelines, by taking an initial guess of the pose, size, and track ID of the boxes, \eg provided by a tracking method or by humans, and letting our method compute the optimally placed annotations.

%% file: 6_acknowledgments.tex
\section*{Acknowledgments}

We acknowledge the support of the Swedish Knowledge Foundation via the industrial doctoral school RELIANT, grant nr: 20220130.